\documentclass{article}
\usepackage{spconf,amsmath,graphicx}
\usepackage{booktabs,arydshln}
\usepackage{hyperref}

\title{Lattice-free MMI adaptation of self-supervised pretrained acoustic models}
\name{Apoorv Vyas$^{\,1\,2}$, Srikanth Madikeri$^{\,1}$, Herv{\'e} Bourlard$^{\,1\,2}$}
\address{
    $^{1}$Idiap Research Institute, Martigny, Switzerland \\
    $^{2}$Ecole Polytechnique F\'ed\'erale de Lausanne, Switzerland\\
    \texttt{\{avyas, msrikanth, bourlard\}@idiap.ch}
}

\begin{document}
\maketitle
\begin{abstract}
In this work, we propose lattice-free MMI (LFMMI) for  supervised adaptation of
self-supervised pretrained acoustic model. We pretrain a Transformer model on
thousand hours of untranscribed Librispeech data followed by supervised
adaptation with LFMMI on three different datasets. Our results show that
fine-tuning with LFMMI, we consistently obtain relative WER improvements of
10\% and 35.3\% on the clean and other test sets of Librispeech (100h), 10.8\%
on Switchboard (300h), and 4.3\% on Swahili (38h) and 4.4\% on Tagalog (84h)
compared to the baseline trained only with supervised data.
\end{abstract}

\section{Introduction}
\label{sec:intro}

A typical approach to exploit unsupervised data for automatic speech
recognition (ASR) is to train a seed model using supervised data and use the
seed model to automatically transcribe the unsupervised
data~\cite{zavaliagkos1998using,wessel2005unsupervised,thomas2013deep,vesely2017semi,novotney2009unsupervised,grezl2013semi,zhang2014semi,manohar2018semi,tong2019unbiassed}.
This approach is referred to as semi-supervised learning.  More recently,
self-supervised learning methods that transform the input signal to learn
powerful representations have been receiving a lot of attention
\cite{baevski2020wav2vec,chung2019unsupervised,liu2020tera,oord2019representation,schneider2019wav2vec,wang2020speechbert}.
In contrast to semi-supervised learning, self-supervised learning aims to
improve the seed model by exploiting unlabelled data before adaptation on
supervised data.

Semi-supervised learning and self-supervised learning are complimentary as
using the self-supervised pretrained network as the staring point for
supervised adaptation can potentially improve the seed model which will in-turn
improve the quality of transcriptions generated for the unlabelled data. Thus,
in this work we focus on adapting a self-supervised pretrained network to get
an improved seed model.

Self-supervised training approaches can be broadly grouped into two classes:
(1) auto-regressive models that try to predict the future representations
conditional on the past inputs
\cite{oord2019representation,chung2019unsupervised} and (2) bidirectional
models that learn to predict masked parts of the input
\cite{baevski2020wav2vec,wang2020speechbert,liu2020tera}. In \cite{liu2020tera,
baevski2020wav2vec}, authors explore the adaptation of a bidirectional network
for ASR. However, they only consider cross-entropy based HMM-DNN systems or
Connectionist Temporal Classification \cite{graves2006ctc} for supervised
training. Moreover, they evaluate the supervised adaptation when unlabelled
data and transcribed data belong to the same domains. 

The contribution of this work is twofold. First, we show that for the same
pretraining method, using LFMMI as the sequence discriminative criterion for
adaptation yields a better performance than cross-entropy training. Second, we
show that the pretraining remains useful even for out-of-domain supervised
data and cross-lingual adaptation.

Specifically, we use the self-supervised pretraining method proposed in
\cite{liu2020tera} to train on thousand hours of librispeech untranscribed
data. We then adapt the pretrained model using LFMMI on three datasets.
First, we evaluate on a hundred hour subset of Librispeech. Next, we train on
the three hundred hours of Switchboard data, which is conversational speech.
Finally, we also evaluate on two low resource languages, Tagalog and Swahili,
from the Babel dataset. We always use flat-start LFMMI \cite{hadian2018flat}
training for ASR.

The rest of the paper is organized as follows: In Section \ref{sec:pretraining}, we
describe the details of self-supervised pretraining such as acoustic features,
input perturbations, and model architectures. In Section \ref{sec:asr}, we present
the details of the data preparation and supervised training for automatic
speech recognition. Finally, in Section \ref{subsec:results}, we present the results
on Librispeech , Switchboard, and Babel.

\section{Self-Supervised Pretraining}
\label{sec:pretraining}
We use the Masked Acoustic Modeling task as described in \cite{liu2020tera} for
self-supervised pretraining of a bidirectional Transformer
\cite{vaswani2017attention} model. The input to the network is a sequence of
acoustic features with a percentage of the input frames masked or noise
corrupted. The model attempts to reconstruct the original input given the
corrupted input and is trained with $\text{L}_1$ loss. 

We use the publicly available Librispeech \cite{panayotov2015librispeech}
dataset that comprises a total of 960 hours of read speech data. It is divided
into three parts: \emph{train-clean-100} (100h), \emph{train-clean-360} (360h),
and \emph{train-other-500} (500h). In contrast to \cite{liu2020tera}, we only
consider 80 dimensional filterbank energy features for pretraining. We do not
use fMLLR features as we need alignments to extract those, which makes them
unsuitable in general unsupervised settings. 

In the following, we briefly describe the input perturbations used for the
Masked Acoustic Modeling task and the model architecture along with the
training details.

\subsection{Input Perturbations}
\label{subsec:perturbations}
As described in \cite{liu2020tera}, we apply the following three perturbations
to the input acoustic features. 

First, we apply time alterations, where we randomly select $T_{num}$ starting
points and then mask $7$ consecutive frames. Time alteration blocks can overlap
each other resulting in altered block with more than $7$ frames. The selected
frames are set to zero with a probability of $0.8$, or replaced with random
segments of frame with a probability of $0.1$. For 10\% of the times, they are
left unaltered. $T_{num}$ is selected such that the total number of masked
frames are roughly $15\%$.

We then apply frequency alteration where we randomly mask a block of
consecutive channels to zero for all time steps across the input sequence. The
number of channels to be masked is selected in the range of 0 to $W_c$ with
equal probability. We set $W_c$ to $16$ in our experiments.

Finally, we apply magnitude alteration where with a probability of $0.15$, we
add a noise matrix of the same shape as acoustic features to our input. Each
element to the noise matrix is sampled from a Gaussian distribution with mean
zero and variance $0.2$.

\subsection{Model Architecture}
\label{subsec:models}
We train two Transformer architectures namely: \emph{Tr-med} and
\emph{Tr-small}. \emph{Tr-med} is composed of 12 encoder layers each with 6
attention heads.  The embedding dimension is set to 64 for each head and
feed-forward dimension is set to 1536.  We use all $960$ hours of data to
pretrain this model.

\emph{Tr-small} has 3 encoder layers; each with 12 attention heads.  We set
embedding dimension to 64 for each head and feed-forward dimension to 3072. We
only use \emph{train-clean-100} subset to pretrain \emph{Tr-small}.

In our experiments, we compare against the pretrained
model presented in \cite{liu2020tera} referred to as \emph{Tr-large}, which has 12
encoder layers. Each layer has 12 attention heads and uses an embedding
dimension of 64 for each head. The feed-forward dimension is set to 3072.
\emph{Tr-large} is also trained with all $960$ hours of data. To handle long
sequences, we use the improved-clustered attention with $200$ clusters as
proposed in \cite{vyas2020fast}.

\subsection{Training}
\label{subsec:mam_pretraining}
Both transformer models are trained with the Adam optimizer
\cite{kingma2014adam} with a mini-batch size of 36. The learning rate is warmed
up over the first 7\% of total training steps $T_{steps}$ to a peak value of $0.0002$
and then linearly decayed. We set $T_{steps}$ to $200\,000$.

\section{Experiments}
\label{sec:asr}

We evaluate the pretrained models in two settings. In the first setting, we freeze the weights of pretrained model and use it as a feature extractor.
We pass the output of the encoder to a twelve layered factorized time-delay
neural network (TDNNF) architecture referred to as \emph{TDNNF-large}. In the
second setting, we fine-tune the pretrained model together with a seven
layered TDNNF architecture \emph{TDNNF-small}. We denote
fine-tuning as FT in later experiments.  For baseline, we train from scratch
\emph{TDNNF-large} using the same 80 dimensional filter bank features which
were used for pretraining. For TDDNF models, we set hidden layer dimension to
1024 and bottleneck dimension to 128. 

For supervised training, we use full biphones to enable flat-start training
\cite{hadian2018flat} with lattice free maximum mutual information (LFMMI).
This enables end-to-end training without the need of prior alignments. We refer
to it as \emph{e2e-lfmmi} in later experiments. Unless specified, we use this
for adaption or training from scratch. We also compare against the cross-entropy based adaptation which is referred to as \emph{hybrid} in later experiments.

All our models are trained in PyTorch
\cite{paszke2019pytorch} using the PkWrap toolkit \cite{madikeri2020pkwrap}. Our training scripts and pretrained model are available for reproducibility.\footnote{\href{https://github.com/idiap/apam}{https://github.com/idiap/apam}}

We apply speed and volume perturbation to increase the dataset to three times.
All our models are trained for 15 epochs with a batch size of $32$. We use Adam
\cite{kingma2014adam} optimizer with a learning rate that is decayed from
$0.001$ to $0.00003$ using a polynomial decay. When fine-tuning the pretrained
model, we set the learning rate for the pretrained network to be $0.00003$ and
use the same learning rate policy for the TDNNF network.

\begin{table*}[th]
    \centering
    \scalebox{0.9}{
    \begin{tabular}{cccccccccc}
    \toprule
      & & &  & & & \multicolumn{4}{c}{\textbf{Word Error Rate}} \\
      \cline{7-10}
      & & &  & & & \multicolumn{2}{c}{\textbf{3-gram}} & 
      	\multicolumn{2}{c}{\textbf{4-gram}} \\
      \cline{7-10}
      & Architecture                             & Features & Pretrain & Supervision & Fine-tune & clean     & other      & clean     & other \\
      \midrule
      \multicolumn{10}{c}{Pretraining + Supervised} \\
      \midrule
      (a) & Tr-small + TDNNF-large               & FBANK    & 100h      & e2e-lfmmi   & No       & 8.91      &  25.33     & 6.09      & 18.80 \\
      (b) & Tr-small + liGRU  \cite{liu2020tera} & FBANK    & 100h      & hybrid      & No       & 11.83     &  NA        & 9.43      & NA \\
      (c) & Tr-med + TDNNF-large                 & FBANK    & 960h      & e2e-lfmmi   & No       & 7.98      & 22.14      & 5.52      & 16.32 \\
      (d) & Tr-med + TDNNF-small                 & FBANK    & 960h      & e2e-lfmmi   & Yes      & \bf{7.78} & \bf{20.19} & \bf{5.35} & \bf{14.75} \\
      (e) & Tr-large + liGRU \cite{liu2020tera}  & fMLLR    & 960h      & hybrid      & Yes      & 8.23      &  NA        & 5.84      & NA \\
      \midrule
      \multicolumn{10}{c}{Supervised Only} \\
      \midrule
      (f) & TDNNF-large                          & FBANK    & -         & e2e-lfmmi   & -        & 8.64      & 26.27      & 5.89      & 20.02 \\
      \bottomrule
    \end{tabular}
    }
    \caption{
        Comparison of word error rates (WER) (in \%) on the clean and other
        parts of the Librispeech test set with and without 4-gram language
        model rescoring. Fine-tuning the pretrained model with LFMMI
        significantly outperforms other baselines. Tr-small, Tr-med, Tr-large
        refers to small, medium, and large Transformers used for pretraining.
        TDNNF-small and TDNNF-large refer to the $7$ and $12$ layered TDNNF
        architectures used for fine-tuning. liGRU refers to light Gated
        Recurrent Unit.  e2e-lfmmi refers to flat-start training with LFMMI and
        hybrid refers to cross-entropy based HMM-DNN models.
    }
    \label{tab:librispeech}
\end{table*}

\subsection{Datasets}
\label{subsec:datasets}
We evaluated the performance of the pretrained model on three different
datasets with increasing order of difficulties.  The first dataset we consider
is the $100$ hour subset of Librispeech \cite{panayotov2015librispeech} called
\emph{train-clean-100}. This is the easiest setting as there is no domain shift
with respect to the pretraining data. The next dataset we used is the
Switchboard \cite{godfrey1992switchboard} with 300 hours of supervision data.
For both Switchboard and Librispeech (pretraining data) the spoken language is
English. However Switchboard has conversational speech while Librispeech is
read speech. We also present results on two of the Babel \cite{gales2014speech}
languages: Tagalog (84.h) and Swahili (38.5h). This is the hardest setting we
consider as there is both language and acoustic conditions mismatch.

\subsection{Results}
\label{subsec:results}
In the following we compare the Word Error Rate (WER) achieved by
supervised adaptation of pretrained models to the models trained from scratch. The number of hours denote the transcribed data for adaptation. Unless specified, we use the pretrained model trained on $960$ hours of Librispeech data.

\subsubsection{Librispeech ($100$ hours)}
\label{subsubsec:libri}

In this experiment, we discuss the setting when the pretrained data and
labelled data for ASR come from the same domain. We present our main results in
the Table \ref{tab:librispeech}. For decoding, we use the model that achieves
lowest WER  on the \emph{dev-clean} set. We first decode using a trigram
language model and then rescore using a $4$-gram language model.

In rows (a) and (b), we compare the effect of training criterion
used for adaptation of the pretrained model. It can be seen that using LFMMI
for adaptation outperforms hybrid models when FBANK features are used for
self-supervised pretraining. Moreover, compared to training from scratch using
only the labelled data (f), using \emph{Tr-small} (a) performs slightly worse
on the clean data.  This is expected because \emph{Tr-small} was pretrained using
the same $100$ hour subset which is used for training (f). Thus both models are
exposed to exactly the same amount of data during training. Interestingly (a)
performs better than (f) on the other portion of the test data. We think that these gains are a result of noise robustness due to the perturbations added during pretraining.

In rows (c) and (d), we present the results of pretraining using entire $960$
hours of data with \emph{Tr-med}. We can see that fine-tuning the pretrained
model leads to a better performance than simply using it as a feature
extractor. Both (c) and (d) outperform the pretrained model with fMLLR
features (e) as well as the training only with labelled data (f). Note that we
get ~6\%  absolute improvements on the other portion of the test set when
fine-tuning with LFMMI. We hypothesize that we get such improvements on the
\emph{other} portion because the pretrained model sees that part of the
dataset during the self-supervised pretraining.

From Table \ref{tab:librispeech}, it can be concluded using LFMMI for
supervision not only results in better performance but also reduces the
sensitivity to the features used.

\subsubsection{Switchboard ($300$ hours)}
\label{subsubsec:swbd}
\begin{table}[th]
    \centering
    \scalebox{0.9}{
    \begin{tabular}{lcccc}
        \toprule
        & \multicolumn{4}{c}{\textbf{Hub5'00 (eval2000)}} \\
        \cline{2-5}
        & \multicolumn{2}{c}{\textbf{3-gram}} & \multicolumn{2}{c}{\textbf{4-gram}} \\
        \textbf{Model} & \textbf{SWBD} & \textbf{CH} & \textbf{SWBD} & \textbf{CH} \\
         \midrule
         \multicolumn{5}{c}{Pretraining + Supervised} \\
         \midrule
         Tr-med + TDNNF-large     & 11.3      & 22.0      & 9.9      & 19.7 \\
         Tr-med + TDNNF-small (FT)& \bf{10.9} & \bf{20.4} & \bf{9.4} & \bf{18.2} \\
         \midrule
         \multicolumn{5}{c}{Supervised Only} \\
         \midrule
         TDNNF-large          & 11.8      & 22.5      & 10.3     & 20.3 \\
         TDNN-LSTM \cite{hadian2018flat}  & 11.3     & 21.5      & 9.8    & 19.3 \\
        \bottomrule
    \end{tabular}
    }
    \caption{
        Comparison of word error rates (WER) (in \%) on eval2000 test set for
        the 300 hours Switchboard task. The 3-gram language model is based on
        Switchboard, whereas the 4-gram employs Switchboard+Fisher training set
        transcripts. Using pretrained model as a feature extractor or
        fine-tuning outperforms the baseline.  Fine-tuning
        achieves the best WER on both callhome and switchboard part of eval
        set.
    }
    \label{tab:swbd}
\end{table}
In this experiment, we explore the case when the pretraining data and labelled
data for ASR belong to the same language but are different with respect to
content, speakers, and acoustic conditions. Switchboard has conversational
speech recorded at $8$ KHz while Librispeech has read speech at $16$ KHz. To be
compatible with the pretrained models, we resample the Switchboard recordings
at $16$ KHz before extracting the features.  For the \emph{TDNNF-large}
baseline trained only with labelled data, we use the $8$ KHz recordings.

Table \ref{tab:swbd} compares the WER for the models trained from scratch to
those that make use of Librispeech pretraining data. In both cases, the
pretrained models outperform the model trained from scratch. Once again, the 
fine-tuned model results in most improvements in terms of WER.

\subsubsection{Babel: Swahili and Tagalog}
\label{subsubsec:babel}
In our final experiment, we consider the scenario when the pretraining data
and labelled data for ASR do not share the same language or the acoustic
conditions. For this task, we consider the two low resource languages from
Babel database: Swahili and Tagalog. Similar to Switchboard setup, we resample
the recordings at $16$ KHz to be compatible with the pretrained model. Once
again, for the \emph{TDNNF-large} model trained from scratch only on the
supervised data, we use the $8$ KHz recordings.

Due to the lack of a separate evaluation set, we report results on the
\emph{dev10h} development part of both languages. We remove $1000$ utterances
from the training set to be used as the development set for model selection. We
use trigram language model for decoding.

Table \ref{tab:babel} compares the WER for the models trained from scratch to
those that make use of Librispeech pretraining data. It can be seen that for
both Swahili and Tagalog, using pretrained model as feature extractor performs
worse than the models trained from scratch. This indicates the representations
learnt by the pretrained model on Librispeech data removes some important
information that are specific to these languages resulting in worse performance
than the baseline. However, on fine-tuning, the model adjusts its parameters to
recapture this information and outperforms the baseline.

\begin{table}[th]
    \centering
    \scalebox{0.9}{
    \begin{tabular}{lcc}
        \toprule
        \textbf{Model} & \textbf{Swahili} & \textbf{Tagalog} \\
         \midrule
         \multicolumn{3}{c}{{Pretraining + Supervised}} \\
         \midrule
         Tr-med + TDNNF-large      & 40.4      &     46.6   \\
         Tr-med + TDNNF-small (FT) & \bf{36.7} & \bf{43.4}  \\
         \midrule
         \multicolumn{3}{c}{{Supervised Only}} \\
         \midrule
         TDNNF-large          & 38.3     & 45.3 \\
         \midrule
         \multicolumn{3}{c}{Baselines from other work} \\
         \midrule
         Multi-10 \cite{inaguma2019transfer} & 41.6 & 46.2 \\
         BLSTM-HMM \cite{inaguma2019transfer} & 38.3 & 46.3 \\
        \bottomrule
    \end{tabular}
    }
    \caption{
        Comparison of word error rates (WER) (in \%) on dev10h set for
        the Swahili and Tagalog languages of the Babel dataset. Fine-tuning
        the pretrained model with Librispeech data significantly outperforms the
        monolingual and other reported baselines.
    }
    \label{tab:babel}
\end{table}

\section{Conclusions and Future Work}
We proposed lattice-free MMI for supervised adaption of a self-supervised
pretrained network for acoustic modeling. We show that fine-tuning a
pretrained model with LFMMI criterion outperforms models that are only trained
with supervised data resulting in an improved seed model. We further show that
pretraining remains useful even under strong distributional shifts. 

In future, we intend to combine self-supervision based approaches with
multi-lingual training and iterative decoding based semi-supervised training
approaches to further improve the performance in low resource settings. We will
also fairly compare the performance of other suggested self-supervised training
approaches such as \emph{wav2vev} when the model capacity and pretraining data
is fixed.

\section{Acknowledgments}
This work has received funding from the
Swiss National Science Foundation project SHISSM (Sparse and
hierarchical Structures for Speech Modeling), grant agreement 200021-175589. The research is also partially based upon the work supported by the Office of
the Director of National Intelligence (ODNI), Intelligence Advanced Research
Projects Activity (IARPA), via AFRL Contract \#FA8650-17-C-9116. The views and
conclusions contained herein are those of the authors and should not be
interpreted as necessarily representing the official policies or endorsements,
either expressed or implied, of the ODNI, IARPA, or the U.S. Government. The
U.S. Government is authorized to reproduce and distribute reprints for
Governmental purposes notwithstanding any copyright annotation thereon.
\bibliographystyle{IEEEbib}
\bibliography{strings,refs}

\end{document}